\documentclass[twoside]{article}

\usepackage[preprint]{aistats2026}
\usepackage[round]{natbib}

\usepackage{amsfonts} 
\usepackage{enumitem}
\usepackage{amsthm}
\usepackage{mathtools}
\usepackage{algpseudocode}
\usepackage{algorithm}
\usepackage{delimset}
\usepackage{hyperref}
\usepackage[capitalize]{cleveref}

\newcommand{\ctg}[1]{\mathcal{V}^{#1}}

\newtheorem{theorem}{Theorem}[section]

\theoremstyle{definition}
\newtheorem{definition}{Definition}

\newtheorem{lemma}{Lemma}
\newtheorem{assumption}{Assumption}
\newtheorem{observation}{Observation}
\newtheorem{corollary}{Corollary}[theorem]

\newcommand{\costbound}{B_\star}
\newcommand{\timebound}{T_\star}

\newcommand{\indgeventi}[1]{\mathbb{I} \{ \Omega^{#1} \}}

\newcommand{\sinit}{s_\text{init}}

\newcommand{\cmdp}[1]{\mathcal{M} ({#1})}
\newcommand{\cmdpk}[1]{\mathcal{M}^{\text{know}} ({#1})}
\newcommand{\cmdpkt}[1]{\widetilde{\mathcal{M}}^{\text{know}} ({#1})}

\newcommand{\highprobcostbound}[1]{48 \costbound \log \frac{4 #1}{\delta}}

\newcommand{\valf}[2]{\mathcal{V}^{\pi}_{\mathcal{M}(#1)} (#2)}
\newcommand{\propset}[1]{\Pi_\text{proper}({#1})}

\newcommand{\optimisticctg}[1]{\widetilde{\mathcal{V}}^{#1}}

\DeclareMathOperator*{\argmin}{arg\,min}

\begin{document}

%

%

\twocolumn[

\aistatstitle{Regret Guarantees for Linear Contextual Stochastic Shortest Path}

\aistatsauthor{ Dor Polikar \And Alon Cohen }

\aistatsaddress{ School of Electrical Engineering, \\ Tel Aviv University \And  School of Electrical Engineering, \\ Tel Aviv University \\ Google Research, Tel Aviv } ]

\begin{abstract}
    We define the problem of linear Contextual Stochastic Shortest Path (CSSP), where at the beginning of each episode, the learner observes an adversarially chosen context that determines the MDP through a fixed but unknown linear function. 
    The learner's objective is to reach a designated goal state with minimal expected cumulative loss, despite having no prior knowledge of the transition dynamics, loss functions, or the mapping from context to MDP.
    In this work, we propose LR-CSSP, an algorithm that achieves a regret bound of $\widetilde{O}(K^{2/3} d^{2/3} |S| |A|^{1/3} \costbound^2 \timebound \log (1/ \delta))$, where $K$ is the number of episodes, $d$ is the context dimension, $S$ and $A$ are the sets of states and actions respectively, $\costbound$ bounds the optimal cumulative loss and $\timebound$, unknown to the learner, bounds the expected time for the optimal policy to reach the goal. In the case where all costs exceed $\ell_{\min}$, LR-CSSP attains a regret of $\widetilde O(\sqrt{K \cdot d^2 |S|^3 |A| \costbound^3 \log(1/\delta)/\ell_{\min}})$.
    Unlike in contextual finite-horizon MDPs, where limited knowledge primarily leads to higher losses and regret, in the CSSP setting, insufficient knowledge can also prolong episodes and may even lead to non-terminating episodes. 
    Our analysis reveals that LR-CSSP effectively handles continuous context spaces, while ensuring all episodes terminate within a reasonable number of time steps.
\end{abstract}

\section{Introduction}
We study the problem of learning Contextual Stochastic Shortest Path (CSSP), where the learner receives a context---additional side information---before each episode, that determines a specific SSP instance the learner faces during the episode. 
In each such instance, the learner attempts to reach a prespecified goal state while minimizing their expected cumulative loss.

Our problem is a special case of Reinforcement learning (RL)---a powerful framework for modeling sequential decision-making problems in unknown and stochastic environment.
These environments are typically modeled as Markov decision processes \citep[MDP]{10.5555/528623} that are assumed to remain fixed throughout the learning process.
However, in many real-world applications---such as personalized medical treatment---the underlying decision-making dynamics may vary between individuals, with each instance effectively representing a different MDP.

Personalized medical treatment can be modeled as stochastic shortest path problem (SSP)---a special type of MDP, where a learner aims to reach a predefined goal state while minimizing their expected total loss. 
For example, consider a patient who has undergone a blood test revealing several abnormal biomarkers. 
The treatment process aims to bring these biomarkers into a healthy range, with each action (e.g., medication, lifestyle change) incurring a cost or risk. 
Importantly, both the transition dynamics and associated losses may differ between patients, reflecting variability in physiology and treatment response. 
The Contextual Stochastic Shortest Path (CSSP) model captures such personalized settings, allowing for individual-specific dynamics and objectives.
More generally, SSP also generalizes both finite-horizon and discounted MDPs, and models a wide range of applications including game playing, autonomous driving, and finetuning of large language models.

Learning a single non-contextual instance of SSP, was originally suggested by~\citet{10.5555/3524938.3525812} who gave an algorithm with a $\widetilde{O}(K^{2/3})$ regret guarantee.
\citet{pmlr-v119-rosenberg20a,cohen2021minimax,tarbouriech2021stochastic} later improved this to a tight bound of $\widetilde{\Theta}(\sqrt{(\costbound+\costbound^2) |S| |A| K})$, where $S$ and $A$ denote the state and action space respectively, $\costbound$ is an upper bound on the expected cumulative loss of the optimal policy from any initial state and $K$ is the number of episodes. 

Contextual RL has also been extensively studied. 
Contextual Multi-Armed Bandits (CMABs) and Contextual finite horizon MDPs (CMDPs) extend classical decision-making models by incorporating context into the learning process. 
CMABs use context to guide action selection, with applications in online advertising and personalized recommendations (see, e.g., \citealp{9185782}). 
CMDPs, originally suggested in \citet{hallak2015contextualmarkovdecisionprocesses}, generalize this approach by modeling state transitions, with context defining distinct MDPs. 
CMDPs were later studied in \citet{pmlr-v202-levy23a, pmlr-v83-modi18a,qianoffline,levy2024eluder}, where the focus was on the context-to-dynamics mapping, and how to use it for computationally-efficient learning.

Our work borrows from \cite{pmlr-v83-modi18a}, and assumes that both the transition dynamics and the loss function decompose into linear context-dependent embeddings. The contexts are assumed to be generated arbitrarily, possibly adversarially.
Our algorithm LR-CSSP is based on the well-established principle of \emph{Optimism in the Face of Uncertainty} (see, e.g., \citealp{NIPS2008_e4a6222c}). 
It maintains confidence sets over the corresponding embeddings, which are guaranteed to contain the true parameters with high probability. 
LR-CSSP is guaranteed to reach the goal state while incurring a total regret of $\widetilde{O}(K^{2/3} d^{2/3} |S| |A|^{1/3} \costbound^2 \timebound \log (1/ \delta))$, where $d$ is the context dimension, $\costbound$ bounds the optimal cumulative loss over all possible SSP instances and $\timebound$, unknown to the learner, bounds the expected time for the optimal policy to reach the goal. In the case where all losses exceed a minimal value $\ell_{\min}$, thus eliminating the risk of getting stuck in a loss-free loop, LR-CSSP attains regret of $\widetilde O(\sqrt{K \cdot d^2 |S|^3 |A| B^3_* \log(1/\delta)/\ell_{\min}})$.

One of the main challenges in learning SSPs, which was also a challenge in previous work, is making sure that each episode terminates by reaching the goal state. 
Previous work mitigates this issue by dichotomizing state-action pairs into known, sufficiently sampled state-action pairs, and unknown. 
Once all state-action pairs are known, due to the optimistic approach, the algorithm is guaranteed to reach the goal state.
Here, because of the contextual nature of the problem, there is an additional challenge of generalizing between different contexts in order to make sure episodes terminate quickly on new, possibly previously unseen, contexts.
We still rely on known and unknown state-action pairs. However, as the contexts arrive arbitrarily, state-actions may fluctuate between known and unknown until they become permanently known.

An important future direction is extending LR - CSSP to settings with non-linear context dependencies. While linear embeddings offer tractability and theoretical guarantees, many real-world systems exhibit complex, non-linear dynamics. Incorporating kernel methods or deep learning-based representations could enhance expressiveness, allowing the model to capture richer structure while preserving sample efficiency through appropriate regularization or oracle-based strategies.

We see our work as a first step in studying general, other than finite horizon, CMDPs. Our hope is that future work tackles more general, non-linear, function approximation, possibly using oracle approaches such as in \cite{pmlr-v202-levy23a}.

\subsection{Additional related work}
There is a rich body of work on regret minimization in RL, much of which is based on the optimism principle. 
The majority of this literature focuses on the tabular setting~\citep{NIPS2008_e4a6222c,pmlr-v70-azar17a,NEURIPS2018_d3b1fb02,fruit2018efficientbiasspanconstrainedexplorationexploitationreinforcement,pmlr-v97-zanette19a,NEURIPS2019_25caef3a,NEURIPS2019_10a5ab2d}. 
Recent advances have extended these ideas to function approximation under various structural assumptions, such as linearity, low-rank dynamics, or realizability \citep{pmlr-v97-yang19b, pmlr-v125-jin20a, pmlr-v108-zanette20a, pmlr-v119-zanette20a, pmlr-v162-vial22a, pmlr-v162-chen22h, pmlr-v162-min22a, pmlr-v202-di23a}.
\citet{pmlr-v162-min22a} study a stochastic shortest path problem with linear function approximation, providing a regret bound that characterizes learning performance under structural assumptions. ~\citet{pmlr-v201-dann23b} proposed unified algorithmic frameworks that address a broad class of stochastic path problems, integrating various formulations such as SSPs and budgeted path planning into a single model with strong theoretical guarantees.

Sample complexity and regret guarantees for Contextual Decision Processes (CDPs) and Contextual MDPs (CMDPs) have been studied under various assumptions. 
OLIVE~\citep{pmlr-v70-jiang17c} provides sample-efficient learning under the low Bellman rank assumption, while Witness Rank \citep{pmlr-v99-sun19a} enables PAC bounds for model-based learning. 
Generalization in smooth CMDPs and linear mixtures of MDPs is addressed by \citet{pmlr-v83-modi18a}, and \citet{pmlr-v202-levy23a} provide a reduction from stochastic CMDPs to supervised learning. 
CMDPs extend Contextual Multi-Armed Bandits (CMAB), for which regret has been well characterized under linear assumptions \citep{DBLP:journals/corr/abs-2007-07876, pmlr-v15-chu11a} and function approximation tools \citep{pmlr-v119-foster20a}. 

\section{Preliminaries}
\subsection{Stochastic Shortest Path (SSP)}
The Stochastic Shortest Path (SSP) is a Markov Decision Process (MDP) problem in which the learner interacts with an MDP $\mathcal{M}=(S, A, P, \ell,\sinit,g)$, where $S$ and $A$ are finite sets of states and actions, respectively. 
The interaction with $\mathcal{M}$ starts in an initial state, at $s_0 = \sinit \in S$. The learner transitions between states according to the dynamics: $P(\cdot \mid s, a)$ and the interaction is completed only when reaching $g \notin S$.
During the course of its interaction, it suffers losses $\ell(s,a) \in [0, 1]$ every step according to $\ell(s,a) \sim \mathcal{D}_{s,a}$ with the objective of minimizing the expected cumulative loss on all steps and reaching $g$. 
The amount of steps required to reach $g$ is, in general, unknown and depends both on $\mathcal{M}$ and the policy $\pi$. 
We denote this quantity $I_k$, which may be infinite.

A \emph{stationary and deterministic policy} $\pi : S \mapsto A$ defines a mapping from states to actions. Given an MDP $\mathcal{M}$, any policy $\pi$ induces a value function of a state $s$ over the horizon $H$ is defined as follows:
\begin{align*}
    \mathcal{V}^{\pi} (s)
    = 
    \mathbb{E}_{\mathcal{M}, \pi} 
    \brk[s]*{\sum_{t=0}^{I_k} \ell(s_t, a_t) \mid s_0=s}
    .
\end{align*}
\paragraph{Proper policies.} This work addresses the approximation of the optimal \textit{proper} policy. Namely, a policy that guarantees reaching the goal state $g$ from any initial state:
\begin{definition}[Proper and Improper Policies]
    A policy $\pi$ is \emph{proper} if playing $\pi$ reaches the goal state with probability $1$ when starting from any state.
    A policy is \emph{improper} if it is not proper.
\end{definition}
Note that the optimal proper policy may not necessarily be the optimal policy overall, in particular when there are loops with zero cumulative loss.

For a proper policy $\pi$, the value function $\mathcal{V}^\pi(s)$ is guaranteed to be finite for all $s \in S$ due to the finiteness of $S$. 
However, $\mathcal{V}^\pi(s)$ may be infinite if $\pi$ is improper. 
Let $\ T^\pi(s) $ the expected time it takes for $\pi$ to reach $g$ starting at $s$. 
In particular, if $\pi$ is proper then $T^{\pi}(s)$ is finite for all $s$; in contrast, if $\pi$ is improper there exist at least one $s$ such that $T^{\pi}(s) = \infty$.

The following Lemmas characterize the importance of proper policies:

\begin{lemma}[\normalfont{\citealp[Lemma 1]{bertsekas1991analysis}}]
    \label{lem:bertsekas-proper}
    Suppose that there exists at least one proper policy and that for every improper policy \(\pi'\) there exists at least one state \(s \in S\) such that \(\ctg{\pi'}(s) = \infty\). 
    Let \(\pi\) be any policy, then
    \begin{enumerate}[nosep,leftmargin=*,label=(\roman*)]
        \item 
        If there exists some \(\ctg{} : S \mapsto \mathbb{R}\) such that 
        \(
            \ctg{}(s) 
            \ge 
            \ell \bigl(s, \pi(s) \bigr) 
            + 
            \sum_{s'\in S} P \bigl(s' \mid s, \pi(s) \bigr) \ctg{}(s')
        \)
        for all \(s \in S\), then \(\pi\) is proper. 
        Moreover, it holds that 
        \(
            \ctg{\pi}(s) \leq \ctg{}(s), \; \forall s \in S.
        \)
        \item 
        If \(\pi\) is proper then \(\ctg{\pi}\) is the unique solution to the equations 
        \(
            \ctg{\pi}(s) 
            = 
            \ell \bigl(s, \pi(s) \bigr) 
            + 
            \sum_{s'\in S} P \bigl(s' \mid s, \pi(s) \bigr) \ctg{\pi}(s')
        \)
        for all \(s \in S\).
    \end{enumerate}
\end{lemma}

\begin{lemma}[\normalfont{\citealp[Proposition 2]{bertsekas1991analysis}}]
    \label{lem:bertsekas-optimal}
Under the conditions of \cref{lem:bertsekas-proper}
the optimal policy \(\pi^\star\) is stationary, deterministic, and proper. Moreover, a policy \(\pi\) is optimal if and only if it satisfies the Bellman optimality equations for all \(s \in S\):
    \begin{alignat}{2} \label{eq:bellman}
        &\ctg{\pi}(s) 
        &&= 
        \min_{a \in A} \ell \bigl(s, a \bigr) + \sum_{s'\in S} P \bigl(s' \mid s, a \bigr) \ctg{\pi}(s'), \\
        &\pi(s) 
        &&\in 
        \argmin_{a \in A} \ell \bigl(s, a \bigr) + \sum_{s'\in S} P \bigl(s' \mid s, a \bigr) \ctg{\pi}(s').
        \nonumber
    \end{alignat}
\end{lemma}

\cref{lem:bertsekas-proper} gives us the conditions for a policy to be proper.~\cref{lem:bertsekas-optimal} restricts the overall optimal policy to be \textit{stationary, deterministic, and proper} and gives conditions for such policies to be optimal.
%
\subsection{Concentration inequality for linear regression}
\label{sec:confidenceEllipsoid}

Let $ \theta^\star \in \mathbb{R}^d $ be an unknown but fixed parameter vector. At each time step $ t = 1, 2, \dots, T $, an input vector $ X_t \in \mathbb{R}^d $ is possibly a random variable that is a function of $Y_1, Y_2, \dots,Y_{t-1}$, and the learner observes:
\[
    Y_t = \langle \theta^\star, X_t \rangle + \eta_t,
\]
where $ \langle \cdot, \cdot \rangle $ denotes the standard inner product in $ \mathbb{R}^d $, $ \eta_t $ is a bounded noise term conditioned on $X_1, X_2, \dots, X_t$ with zero mean. Let $\hat{\theta}_t$ denote the ridge regression of $\theta^\star$ with regularization parameter \(\lambda>0\). Then, the estimate is given by:
\begin{equation}
    \hat{\theta}_t=(X_{1:t}^TX_{1:t}+\lambda I)^{-1}X_{1:t}^TY_{1:t}.
\end{equation}
where $X_{1:t} \in \mathbb{R}^{t \times d}$ is the matrix whose rows are the vectors $X_1, X_2, \ldots, X_t$, and $Y_{1:t} \in \mathbb{R}^t$ is the vector of corresponding observations $Y_1, Y_2, \ldots, Y_t$.

\begin{theorem}\label{thm:confidenceEllipsoid}
(Paraphrased from~\citealp{NIPS2011_e1d5be1c}).
Let \(\{F_t\}_{t=0}^\infty\) be a filtration. 
Let \(\{\eta_t\}_{t=1}^\infty\) be a a zero mean variable bounded  by $R$ such that each \(\eta_t\) is \(F_t\)-measurable.

Let \(\{X_t\}_{t=1}^{\infty}\) be an \(\mathbb{R}
^d\)-valued stochastic process such that \(X_t\) is \(F_{t-1}\) measurable
and let \(\lambda>0\)  and for any \(t\geq0\) define:
$
    \bar{V}_t = \lambda I + \sum_{s=1}^{t}X_sX_s^T,
    S_t=\sum_{s=1}^t \eta_sX_s.
$

Define \(Y_t = \langle X_t, \theta_* \rangle + \eta_t\),
and assume that \(\|\theta_*\|_2 \leq \beta\) and that for all \(t \geq 1\), \(\| X_t\|_2 \leq L\). Then, for any \(\delta>0\), with probability at least \(1-\delta\), for all \(t\geq0\), \(\theta_*\) lies in the set:

\[
    \begin{aligned}[t]
        \Theta_t = \Bigl\{ \theta \in \mathbb{R}^d :
        \|\hat{\theta}_t - \theta\|_{\bar{V}_t}
        &\le 
        R \sqrt{d \log\frac{1 + t L^2/\lambda}{\delta}} \\
        & + \sqrt{\lambda}\,\beta
        \Bigr\}.
    \end{aligned}
\]
\end{theorem}

\section{Problem Setup: Adversarial Linear-Contextual SSP}
The Contextual Stochastic Shortest Path (CSSP) problem is a generalization of the SSP and the CMDP models, and is defined by the tuple \((\mathcal{C}, S, A,\mathcal{M})\), where \(\mathcal{C}\), \(S\), \(A\) are the context, state and action spaces, respectively. Unlike \(S\) and \(A\), which are finite and discrete spaces, \(\mathcal{C}\) may be uncountably infinite. The mapping \(\mathcal{M}\)  maps a context \(c\in \mathcal{C}\) to an SSP
\(
    \cmdp{c} 
    =
    (S, A, P^c, \ell^c, s^c_\text{init}, g)
\).

In every episode the learner receives a context $c \in C$, begins the interaction with the SSP \(\cmdp{c}\) in \(s^c_\text{init}\) and ends its interaction with \(\mathcal{M}\) by arriving at the goal state \(g\). Note the \(g\) is fixed across all \(c \in C\) and \(g \notin S\). 
Whenever the learner plays action \(a\) in state \(s\), it incurs a loss \(\ell^c(s,a) \in [0,1]\) and the next state \(s' \in S \cup \{g\}\) is chosen with probability \(P^c(s' \mid s,a)\).
Crucially, both the loss function and the dynamics are context-dependent, making the optimal policy also context-dependent.

Similarly to the non-contextual SSP, we avoid explicitly addressing the goal state \(g\). We assume that the probability of reaching the goal state by playing action \(a\) at state \(s\) is \(P^c(g \mid s,a) =1-\sum_{s'\in S}P^c(s'\mid s,a)\).

\paragraph{Interaction protocol.} The interaction between the learner and the environment is defined as follows.
In each episode \(k=1,2,...,K\):
\begin{enumerate}[nosep,leftmargin=*]
    \item An adversary chooses a context $c_k \in C$.
    
    \item The learner chooses an initial policy $\pi^{c_k}$.
    
    \item The learner observes the trajectory generated by playing $\pi^{c_k}$  in $\mathcal{M}$ until it reaches $g$ during $I_K$ steps. The learner may change its policy at any time-step.       
    \item The learner observes the trajectory from $s^c_\text{init}$ to $g$ denoted as: 
    \begin{align*}
        \sigma^k = (c_k, s_{k, 0}, a_{k, 0}, \ell_{k,0}, &\ldots, 
        \\
        &s_{k, I_{k-1}}, a_{k, I_{k-1}}, \ell_{I_{k-1}}, g).
    \end{align*}
\end{enumerate}
%
%
We adopt the following assumptions to meet the assumptions of 
\cref{lem:bertsekas-proper,lem:bertsekas-optimal}:

\begin{assumption}
    \label{ass:ex-prop}
    For every \(c \in C\) there exists at least one proper policy.
\end{assumption}

\begin{assumption}
    \label{ass:strictly-positive}
    All losses are strictly positive: there exists some \(\ell_{\min} > 0\) such that \(\ell(s,a) \geq \ell_{\min}\) for all $(s,a) \in S \times A$.
\end{assumption}

Note that~\cref{ass:strictly-positive} can be relaxed (\cref{sec:alg-n-rest}).

In addition to these assumptions, our main model-assumption is that both the loss function and the dynamics depend \emph{linearly} on the context:

\begin{assumption}\label{ass:linearDNL}
A CSSP \((\mathcal{C},S, A,\mathcal{M})\) is called linear, if there exists \(d\) such that 
$C = \{ c \in \mathbb{R}^{d} \mid c_i \geq 0 \; \forall i=1,\dots,d,\ \text{and}\  \sum_{i=1}^{d} c_i = 1 \}$ and for every context \(c \in \mathcal{C}\) the associated SSP
\(
    \cmdp{c}
    =
    (S, A, P^c, \ell^c,s^c_\text{init})
\)
can be decomposed as follows:
\begin{enumerate}[nosep,leftmargin=*,label=(\roman*)]
    \item 
    \(
        \mathbb{E}\brk[s]*{\ell^c(s,a)}
        = 
        \sum_{j=1}^{d} c_j \cdot \ell^\star_j(s,a) 
        =
        \langle c,\; L^\star(s,a) \rangle. 
    \) for all $s, a \in S \times A$
    \item 
    For any $s,a,s' \in S \times A \times S$:
    \(
        P^c(s' \mid s,a)
        =
        \sum_{j=1}^{d} c_j \cdot p^\star_j(s' \mid s,a)
        =
        \langle c,\; P^\star(s' \mid s,a) \rangle,
    \)
    where each $p^\star_j(s' \mid s,a)$ defines a legal transition distribution. 
\end{enumerate}
In the above
    \(
        L^\star(s, a) \in \mathbb{R}^d, \; \forall (s, a) \in S \times A
    \),
    and
    \(
        P^\star(s'\mid s,a) \in \mathbb{R}^d, \; (s,a, s')\in S \times A \times S
    \)
are fixed but unknown. 
%
We will sometimes write $P^\star(s,a) \in \mathbb{R}^{|S| \times d}$, referring to a matrix whose rows correspond to $P^\star(s' \mid s,a)$, for all $s' \in S$.
\end{assumption}
%
%
\paragraph{Learning formulation.}
The algorithm's performance is measured by the learner's \emph{regret} over \(K\) episodes, that is, the sum of \(K\) differences between the total loss over the \(I_k\) steps during episode \(k\) and the expected loss of the optimal proper policy corresponding to the context \(c_k\): 
\[
    R_k 
    = 
    \sum_{k=1}^K \bigg[ \sum_{t=1}^{I_k} \ell^{c_k}(s_{k, t}, a_{k, t}) 
    -
    \min_{\pi \in \propset{c_k}} \valf{c_k}{s} \bigg],
\]


where \(\propset{c_k}\) is the set of all stationary, deterministic, and proper policies corresponding to \(\cmdp{c_k}\) (that is not empty by \cref{ass:ex-prop}), and \((s_{k, t}, a_{k, t})\) is the \(t\)-th state-action pair at episode \(k\). 
In the case where \(I_k\) is infinite for some \(k\), we define \(R_k = \infty\).

We denote the optimal proper policy for $\cmdp{c}$ by $\pi_c^\star$, defined as: \[\forall s \in S: \mathcal{V}^{\pi^\star}_c(s) = \argmin_{\pi \in \propset{c}} \mathcal{V}^\pi_{\cmdp{c}}(s).\] 
Let $\costbound > 0$ be an upper bound on the values of $\mathcal{V}_c^{\pi^\star}$ , i.e., $\costbound \geq \max_{s \in S, c \in C} \mathcal{V}_c^{\pi^\star} (s)$ and assume that $\costbound \ge 1$.

We also define $T^{\pi}_{\cmdp{c}}(s)$ the expected time it takes for $\pi$ to reach $g$ starting at $s$ under context $c$. Let $\timebound > 0$ be an upper bound on the times \(T^{\pi^\star_c}_{\cmdp{c}}(s)\), i.e., $\timebound \geq \max_{s \in S, c \in C} T^{\pi^\star_c}_{\cmdp{c}}(s)$.

\section{Algorithm and Main Result}\label{sec:alg-n-rest}

We present the \textit{Linear-Regression Contextual Stochastic Shortest Path} (LR-CSSP, ~\cref{alg:alg}). 
Our approach follows the well-known principle of optimism in the face of uncertainty: it maintains confidence sets that, with high probability, contain the true embedding of both the dynamics $P^\star$ and the loss $L^\star$. 
Each time the learner visits a state classified as ``unknown'' (described below), the algorithm selects optimistic embeddings from these sets by solving the following convex equations:
\begin{alignat}{2}\label{eq:loss-ls}
    &\hat{\ell}_m(s,a) 
    &&=
    \argmin_{L \in \mathbb{R}^d}
        \sum_{(c,\ell)\in D(s,a)} 
            (c_m^\top L - \ell)^2 
            +
            \lambda \|L\|^2, \\
    &\hat{P}_m(s,a)' 
    &&=
    \argmin_{P \in \mathbb{R}^{|S| \times d}}
       \sum_{(s',c)\in D(s,a)} \|Pc - e_{s'}\|_2^2 
        +
        \lambda \|P\|_F^2.
    \label{eq:dynamics-ls}
\end{alignat}
Then, $\hat{P}_m(s,a)$ is derived by projecting $\hat{P}_m(s,a)'$ on the set of all stochastic matrices:
\begin{equation}\label{eq:dynamics-proj}
    \hat P_m(s_,a)
    =
    \argmin_{P \in \Psi} \|P - \hat{P}_m'(s,a))\|_{\bar V^{\tau_{s^m_h,a_h^m}}_{s^m_h,a_h^m}}.
\end{equation}
Here $\|P\|_{V}^2 = \mathrm{tr}(P V P^\top)$, for any $P \in \mathbb{R}^{|S| \times d}$ and $V \in \mathbb{R}^{d \times d}$.
Finally,~\cref{alg:alg} maps $\hat{\ell}_m(s,a)$ and $\hat{P}_m(s,a)$ to the context-specific dynamics and loss via the linear function approximation, and computes an optimistic policy in~\cref{ln:optimism}. Let us denote $\bar V_m(s,a) \coloneq \bar{V}^{\tau}_{s,a}$ where $\tau$ is the number of visits $(s,a)$ had at the beginning of interval $m$. 
Formally, we define a state-action pair $(s, a)$ as \emph{known} in interval $m$, if:
%
%
\begin{align}\label{eq:ineq-known}
    \| c_m \|_{\bar V_m(s,a)^{-1}} 
    <
    \frac{\ell_{\min}}{10 \, \costbound \, \max\Big\{\beta, \sqrt{\log (4 m / \delta)} \Big\}},
\end{align}
where
$
    \beta \coloneqq |S| ( \sqrt{ d \log \frac{8 |S|^2 |A| (1 + \tau_{s,a}/\lambda)}{\delta} } + \sqrt{\lambda}).
$
Unlike in previous work on learning non-contextual SSP, in CSSPs a state-action pair is initially classified as unknown and may fluctuate between being known and unknown multiple times, until it is permanently known, as proved in~\cref{lem:bound-2B-known}. 
If all state-actions are known, the algorithm is guaranteed to reach the goal state in finite time as proved in~\cref{lem:close-mdps-proper} of~\cref{sec:t-bound}.

To ensure that the algorithm reliably reaches the goal state, each episode is divided into intervals. A new interval begins whenever the learner encounters an unknown state or reaches the goal state. At the start of each interval, new estimates of the environment’s dynamics and associated loss are computed (\cref{ln:loss-dynamics-ls}). The estimated dynamics are then projected onto the set of stochastic matrices to ensure it represents a valid probability distribution in~\cref{ln:dynamics-proj}.~\cref{eq:loss-ls,eq:dynamics-ls,eq:dynamics-proj} can be solved efficiently using standard convex optimization techniques. Based on these projected dynamics and the estimated loss, a new optimistic policy is computed for the subsequent interval.

The following theorem bounds the regret of \cref{alg:alg}.

\begin{theorem}
\label{thm:regret-bound-main}
    Set $\lambda = 1$. 
    With probability at least \(1-\delta\) the regret of \cref{alg:alg} is bounded by:
    \begin{align*}
        \widetilde{O}
        \Bigg(
            \frac{d^{2} |A| \costbound^3 |S|^3}{\ell_{\min}^2}
            \log\frac{1}{\delta}
            +
            \sqrt{K \cdot\frac{d^2\vert S\vert^3 \vert A\vert \costbound^3 \log{\frac{1}{\delta}}}{\ell_{\min}} }
        \Bigg).  
    \end{align*}
\end{theorem}

\paragraph{Prior Knowledge of $\ell_{\min}$ and $\costbound$. } Assumption~\ref{ass:strictly-positive} can be circumvented by adding a small fixed loss to the loss of all state-action pairs. After this perturbation, the regret bounds from~\cref{thm:regret-bound-main} continue to hold with $\ell_{\min} \leftarrow \epsilon$, $B_{\star} \leftarrow B_{\star} + \epsilon \timebound$ and an additional regret term $\epsilon K \timebound$. By selecting $\epsilon$ to balance these terms, we obtain the following corollary:
\begin{corollary}
\label{crlry:arbitrary-cost}
Running~\cref{alg:alg} using losses $\ell_\epsilon=\max{(\ell(s,a), \epsilon)}$ for $\epsilon=|S|\sqrt[3]{\frac{d^2|A|}{K}}$ gives the
following regret bound with probability at least $1-\delta$:
\begin{align*}
    R_k
    &= 
    \widetilde{O}
    \Bigg(
        K^{2/3} d^{2/3} |S| |A|^{1/3} 
        \costbound^2 \timebound
        \log\frac{1}{\delta}
    \Bigg).  
\end{align*}  
\end{corollary}
The requirement of knowing $\costbound$ in advance can also be easily circumvented via the standard doubling trick. We initialize $\costbound$ to 1, and whenever the optimistic value function is observed to exceed this threshold, we reset the algorithm and update $\costbound$ to twice its previous value. This procedure increases the regret bound by only a factor of $\log \costbound$.
\paragraph{Notations.} Let \(
D = \{(c_t, s_t, a_t, s_{t+1}, \ell_t)\mid t<T_{tot}(K)\}
\) denote the set of all tuples collected over the $K$ episodes, with $T_{tot}(K)=\sum_{k=1}^{K}I_k$ the total number of steps. We discard episode associations, treating all steps as a single time-ordered sequence. The context $c_t$ remains fixed within each episode and changes only between episodes.
For any state-action pair \((s,a)\), define the subset:
$
D(s,a) = \{(c_t, s_t, a_t, s_{t+1}, \ell_t) \in D \mid (s_t, a_t) = (s,a)\}.
$
\begin{algorithm}[H]
    \caption{\sc Linear Regression for Contextual Stochastic Shortest Path (LR-CSSP)}
    \label{alg:alg}
    \begin{algorithmic}[1]
        \State {\bfseries input:} state space $S$, action space $A$, context space $\mathcal{C}$, initial state $\sinit$, confidence parameter $\delta$, regularization parameter $\lambda$.
        
        \State {\bfseries initialization:} \(\forall{(s,a)}: \tau_{s,a}\gets 0 \) count of visits for (s,a),
        \(\bar{V}^{\tau_{s,a}}_{s,a}\gets \lambda I\); \(D=\emptyset\); an arbitrary policy \(\tilde{\pi}\);
        \(h \gets 1\); \(m \gets 1\).
        
        \State \text{set} \(c_m \in C\) \text{as the current context}.
        \For{\(k=1,2,\ldots\)}
        \State \(s_h^m = \sinit\).

        \While{\(s^m_h \neq g\)}
        
        \State choose \(a_h^m=\hat\pi(s^m_h)\),
        \State observe loss \(\ell_h \sim \ell^{c_m}(s_h^m, a_h^m)\)
        \State and the next state: \(s^m_{h+1} \sim P^{c_m}(\cdot \mid s^m_h,a_h^m)\).
        \State \text{set} \(\tau_{s^m_h,a_h^m} \gets \tau_{s^m_h,a_h^m} +1\).
        \State \text{set} \( \bar V^{\tau_{s^m_h,a_h^m}}_{s^m_h,a_h^m} \gets  \bar V^{\tau_{s^m_h,a_h^m}}_{s^m_h,a_h^m} + c_m c_m^\top\).
        
        \State \text{Insert}: \(D \gets (c_m, s^m_h, a_h^m, s^m_{h+1}, \ell_h)\)
        \If{\text{\cref{eq:ineq-known} is true} \! or \! \(s^m_{h+1} \!=\! g\)}
            \State \# start a new interval
            \If{\(s^m_{h+1} \!=\! g\)}
                \State \text{set} $c_{m+1} \gets c_{k+1}$
            \Else
                \State $c_{m+1} \gets c_m$            
            \EndIf
            \State \(m \gets m+1\), \(h \gets 1\).
            \For{\((s,a) \in S \times A\)}
            \State\label{ln:loss-dynamics-ls} \parbox[t]{0.65\linewidth} {Compute $\hat{\ell}_m(s,a)$ and $\hat{P}_m(s,a)'$ by solving~\cref{eq:loss-ls,eq:dynamics-ls}.}


            \State\label{ln:dynamics-proj} \parbox[t]{0.65\linewidth} {Derive $\hat{P}_m(s,a)$ by projecting $\hat{P}_m(s,a)'$ on the set of all stochastic matrices via~\cref{eq:dynamics-proj}}:
            \EndFor
         \State\label{ln:optimistic-po} \parbox[t]{0.70\linewidth} {Compute optimistic policy as 
         \begin{align*}
            (\pi_m, &\tilde \ell_m,\widetilde P_m) \in
            \\
            &\argmin_{\pi, \ell \in \Theta^\ell_m(c), P \in \Theta^P_m(c)} \mathcal{V}^\pi_{\mathcal{M}(\ell,P)} (s),
         \end{align*} \label{ln:optimism}
         where $\mathcal{M}(\ell,P)$ is the instance of SSP  defined with losses $\ell$ and 
        dynamics $P$, and $\Theta^\ell_m(c), \Theta^P_m(c)$ are 
          defined in \cref{eq:loss-confidence-set-per-context} and~\cref{eq:dynamics-confidence-set-per-context} 
          respectively.}
        \EndIf
        \State \text{set} \(h \gets h+1\).
        \EndWhile
        \EndFor
    \end{algorithmic}
\end{algorithm}
\paragraph{Computational complexity.}
The optimistic optimal policy in~\cref{ln:optimistic-po} of~\cref{alg:alg} can be computed efficiently \cite[see][]{10.5555/3524938.3525812}. 
The core idea is to construct an augmented MDP in which the state space remains unchanged, but the action space is expanded to include tuples of the form $(a, \widetilde{P}, \widetilde{\ell})$ where $a \in A$, $\widetilde{P}$ and $\widetilde{\ell}$ are any transition and loss functions within the confidence sets $\Theta^P_m(c), \Theta^\ell_m(s,a)$ as defined in~\cref{eq:loss-confidence-set-per-context,eq:dynamics-confidence-set-per-context}. It can be shown that the optimistic policy along with the associated optimistic model---those that minimize the expected total cost over all policies and feasible transition functions---correspond to the optimal policy of the augmented MDP. This policy can be computed efficiently using Extended Value Iteration (EVI). Moreover, the updates in \cref{ln:loss-dynamics-ls,ln:dynamics-proj} reduce to convex optimization problems, which can be solved in polynomial time using standard techniques. Consequently, the overall algorithm admits a polynomial-time implementation.

\subsection{Regret analysis}\label{sec:regret-analysis}

We now sketch the proof of \cref{thm:regret-bound-main}, and defer the full details to the supplementary material. The proof begins with the definition of the high probability event (HPE). Under the HPE, the remaining proof is reduced to a deterministic bound on the regret.

\paragraph{High probability event.}
We let $\Omega$ denote the event that the following holds:
\begin{enumerate}[nosep,leftmargin=*]
    \item
        \(\forall (s,a) \in S\times A\), for all \(m \ge 0\), \(\vec{L}^\star(s,a)\) lies in the set:
        \begin{equation}   \begin{aligned}\label{eq:loss-conf-set}
            \Theta_m^\ell(s,a)
            &=
            \Bigg\{ L \in \mathbb{R}^d \;\Big|\;
            \| L(s,a) - \hat{L}_m(s,a) \|_{\bar V_m(s,a)} 
        \\
        &\le 
        \sqrt{\, d \, \log \frac{8 |S| |A| (1+\tau_{s,a})/\lambda}{\delta} \,} 
        + \sqrt{\lambda}
        \Bigg\}.
        \end{aligned}
        \end{equation}

        where $\tau_{s,a}$ here is the number of visits to $s,a$ at the start of interval $m$.
    \item
        \(\forall (s,a) \in S\times A\), for all \(m \ge 0\), \(\vec{P}^\star(\cdot \mid s,a)\) lies in the set:
        \begin{equation}
            \begin{aligned}\label{eq:dynamics-conf-set}
                \Theta_m^P(s,a) 
                &= 
                \Bigg\{P\in\mathbb{R}^{ |S| \times d} ~\bigg|~
                \|P - \hat{P}_m(s,a)\|_{\bar V_m(s,a)}
                \\
                &\le
                \sqrt{d\log{\frac{8|S|^2 |A|(1+\tau_{s,a}/\lambda)}{\delta}}}+\sqrt{\lambda}\Bigg\}.
            \end{aligned}
        \end{equation}
    \item Let $H^m$ be the length of interval $m$. 
    \[
        \sum_{h=1}^{H_m} \ell(s_h^m,a_h^m) \le \highprobcostbound{m}.
    \]
    for all intervals $m$ simultaneously. \label{item:interval-cost-bound}
    
    \item For every interval $m$, define $\widetilde V_m$ as the value function of policy $\pi_m$ over the optimistic model $\mathcal{M}(\tilde \ell_m, \widetilde P_m)$ associated with the context $c_m$.
    Then:

    \begin{align}
        \sum_{m=1}^M
        \bigg(
            \sum_{h=1}^{H^m} &\widetilde V_m(s_{h+1}) 
            - \sum_{s' \in S} P^{c_k}(s' \mid s_h^m,a_h^m)\,
          \widetilde V_m(s')
        \bigg) \nonumber \\
        &\leq \costbound \sqrt{T \log \tfrac{4T}{\delta}}.
        \label{eq:azumma-hpe-main}
    \end{align}
\end{enumerate}

\cref{eq:loss-conf-set,eq:dynamics-conf-set} guarantee that the confidence sets contain the true embeddings. 
\cref{eq:azumma-hpe-main} allows to bound the difference between the realized and expected cumulative cost of the algorithm.

We also define the derived confidence sets for every context $c$ as:
%
%
\begin{equation}\label{eq:loss-confidence-set-per-context}
\begin{aligned}
&\Theta^\ell_m(c) = 
\bigl\{ \ell : S \times A \to \mathbb{R} \;\bigm|\; \\
&\quad \ell(s,a) = L \cdot c, \;
L \in \Theta^\ell_m(s,a),\; \forall (s,a) \in S \times A \bigr\},
\end{aligned}
\end{equation}
\begin{equation}\label{eq:dynamics-confidence-set-per-context}
\begin{aligned}
&\Theta^P_m(c) = 
\bigl\{ P' : S \times A \to \Delta(S) \;\bigm|\; \\
&\quad P'(\cdot \mid s,a) = Pc, \;
P \in \Theta^P_m(s,a),\; \forall (s,a) \in S \times A \bigr\}.
\end{aligned}
\end{equation}
The following lemma ensures that the HPE holds with high probability (proved in~\cref{lem:hpe-interval-loss}):
\begin{lemma}\label{lem:hpe-interval-loss-main}
    The HPE holds with probability at least $1-\delta$.
\end{lemma}
The remainder of the proof bounds the regret deterministically, conditioned on the event $\Omega$. We divide the proof into two parts: the first focuses on bounding the overall number of steps $T$, and the second focuses on bounding the regret.

First, we use the principle of optimism to show that the value function induced by the optimistic policy in the optimistic model (\cref{ln:optimism}) is upper bounded by $\costbound$ (\cref{lem:opt-val-bound}).
Then, we prove that the policies generated by the algorithm are guaranteed to be proper if all state-action pairs are known (\cref{lem:known-state-dynamics,lem:known-state-loss,lem:close-mdps-proper}). 
Indeed, when all state-action pairs are known, the true and the optimistic model are ``close'' enough, so \cref{lem:bertsekas-proper} implies that any proper policy in the optimistic model is also proper in the true model. However, it is important to note that not all state-action pairs may be visited. 
In such cases, the learner may still reach the goal state $g$, but properness can no longer be guaranteed. Nevertheless, we bound the number of such occurrences in~\cref{lem:bound-2B-known}, ensuring that almost all episodes terminate at the goal state.
\cref{ln:dynamics-proj} projects the dynamics embedding on the set $\Psi$ of $|S| \times d$ matrices with stochastic columns. 
As $C$ is defined as the probability simplex in $\mathbb{R}^d$, the projection ensures that $\widehat P_m(s,a) \cdot c$ is indeed a probability distribution over $S$ for any $c \in C$.
Moreover, since $\Psi$ is convex, the projection preserves the statistical guarantee attained by the least squares estimate, as in the following Lemma.
\begin{lemma}\label{lem:dynamics-proj-main}
    \begin{align*}
        \|P^\star(s,a)&-\hat P_m(s,a)\|_{\bar V_m(s,a)}^2
        \le
        \\
        &|S| \brk*{\sqrt{d\log{\frac{8|S|^2 |A|(1+\tau_{s,a}/\lambda)}{\delta}}} +\sqrt{\lambda}}^2.
    \end{align*}
\end{lemma}

\begin{proof}
    From \cref{lemma:dynamics-bound} in~\cref{sec:t-bound} we have with probability at least $1-\delta/8$:
    \begin{align*}
        \|P^\star(s,a)&-\hat{P}_m(s,a)'\|_{\bar V_m(s,a)}^2
        \leq
        \\
        &|S|\Bigg({\sqrt{d\log{\frac{8|S|^2|A|(1+\tau_{s,a}/\lambda)}{\delta}}} +\sqrt{\lambda}} \Bigg)^2.
    \end{align*}
    The next step in \cref{alg:alg},~\cref{ln:dynamics-proj}, is to project \(\hat{P}_m(s,a)'\) on the set \(\Psi\) of matrices with stochastic columns, which contains $P^\star$.
    Therefore:
    $
        \|P^\star(s,a)-\hat{P}_m(s,a)'\|_{\bar V_m(s,a)}^2
        \le
        \|P^\star(s,a) - \hat{P}_m(s,a))\|_{\bar V_m(s,a)}^2.
    $
\end{proof}

Next,~\cref{lem:bound-2B-known} bounds the number of time steps required for a state to become known. We include the full proof below:
\begin{lemma}\label{lem:bound-2B-known}
    Let \(N^+_{s,a}\) denote the number of times 
    the pair (s,a) is unknown. Then:
    \begin{align*}
        N^+_{s, a}
        &\leq
        \frac{8d \costbound^2 |S|^2}{\ell_{\min}^2}\log{(1+\frac{\tau_{s,a}}{\lambda d})} \cdot
        \\
        &\bigg
        (\sqrt{d\log(8|S|^2 |A|\brk*{\frac{1+\tau_{s,a}/\lambda} {\delta}}} + \sqrt{\lambda}
        \bigg)^2.
    \end{align*}
\end{lemma}

 \begin{proof} 
    Denote: 
    \[
        a \coloneqq
        \frac{\ell_{\min}}
        {4\,\costbound\,|S|\Big(
        \sqrt{\,d\log\!\tfrac{8|S|^2|A|(1+\tau_{s,a}/\lambda)}{\delta}\,}
        + \sqrt{\lambda}
        \Big)}.
    \]

    It follows that:
    \(
        a^2N^+_{s,a} 
        < 
        \sum_{m=1}^{M} n_m(s,a) \| c_m\|^2_{\bar{V}_m^{-1}} 
        \le 
        2d\log{(1+\frac{\tau_{s,a}}{\lambda d})},
    \)
    where the right-hand inequality is from \cref{lem:aux-bound-2B-known} in~\cref{sec:t-bound}.
    Finally, isolating $N^+_{s,a}$ gives:
    \(
        N^+_{s,a} < \frac{2 d}{a^2} \log \left(1 + \frac{\tau_{s,a}}{\lambda d}\right).
    \)
\end{proof}
%
%
By examining the algorithm, we obtain the following bound on the number of intervals:
\begin{observation}\label{obsv:num-intervals}
    The number of intervals, M, is bounded as:
    $
        M \leq K + |S||A|N^+
    $
    where \(N^+\) is a bound on the number of visits it takes for any $(s,a)$ to be known.
\end{observation}

Now we are ready to state a key lemma of our analysis, a bound on the time it takes to complete $K$ episodes. 
The proof is a combination of \cref{lem:bound-2B-known}, Observation~\ref{obsv:num-intervals} and \cref{item:interval-cost-bound} of the HPE (full proof in~\cref{proof:number-steps-bound}).

\begin{lemma}\label{lemma:number-steps-bound-main}
    Suppose the HPE holds. Then the total number of steps of the algorithm is:
    \begin{align}
       T
       =
        \widetilde{O}
        \Bigg(
            \frac{K \costbound}{\ell_{\min}} + \frac{|S|^3|A|{d^{2} B^3_*}}{\ell_{\min}^3}\log^{2}{\frac{1}{\delta}}
        \Bigg).           
    \end{align}
\end{lemma}

We move now to bound the regret. Our regret analysis starts with the following regret decomposition:
\begin{align}
    R_K
    \label{eq:hoff-reg-decomp-1-main}
    &= \sum_{m=1}^M \sum_{h=1}^{H^m} 
    \left(
        \widetilde V_m (s_{h^m}) - \widetilde V_m (s_{h+1}^m)
    \right)
    \\ \nonumber
    &\qquad \qquad- \sum_{k=1}^K \min_{\pi \in \propset{c_k}} \widetilde V_m(\sinit)
    \\  
    \label{eq:hoff-reg-decomp-2-main}
    &+
    \sum_{m=1}^M  \sum_{h=1}^{H^m}  \sum_{s' \in S}
    \widetilde V_m (s'_m) \cdot
    \\ \nonumber
    &\qquad \qquad 
    \bigg(
        P^{c_k}(s' \mid s_h^m,a_h^m) -
        \widetilde{P}^{c_k}_m(s' \mid s_h^m,a_h^m) 
    \bigg)
\end{align}
\begin{align}
    \label{eq:hoff-reg-decomp-3-main}
    &+ 
    \sum_{m=1}^M
    \bigg( \sum_{h=1}^{H^m}
        \widetilde V_m (s_{h+1}^m)
        \\  \nonumber
        &\qquad \qquad-
        \sum_{s' \in S} P^{c_k}(s' \mid s_h^m,a_h^m)
        \widetilde V_m (s'_m)
    \bigg).
    \\
    \label{eq:hoff-reg-decomp-4-main}
    &+ 
    \sum_{m=1}^M \sum_{h=1}^{H^m} \ell^{c_k}(s,a) - \widetilde{\ell}^{c_k}(s,a).
\end{align}

\cref{eq:hoff-reg-decomp-1-main} captures the loss incurred from switching policies whenever an unknown state-action pair or the goal state are encountered.  


\cref{eq:hoff-reg-decomp-2-main} bounds the transition model misspecification weighted by the true value function. The proof leverages concentration inequalities applied to the empirical transition estimates, along with the definition of the confidence sets. It closely parallels the proof for the error in estimating the cost function, which we provide below for completeness.

\begin{lemma}
\label{lem:sum-conf-bounds-main}
Suppose that $\Omega$ holds. It holds that:
\begin{align*}    
    \sum_{m=1}^M  &\sum_{h=1}^{H^m}  \sum_{s' \in S} \widetilde{V}_m (s'_m) 
    \bigg(
        P^{c_k}(s' \mid s_h^m,a_h^m) 
        \\
        &\qquad \qquad-  \widetilde{P}_m^{c_k}(s' \mid s_h^m,a_h^m) 
    \bigg) \\
    & \leq 
    \costbound \vert S \vert \Bigg(\sqrt{d\log{\frac{8|S|^2 |A| (1+T/\lambda)}{\delta}}} +\sqrt{\lambda}\Bigg) \cdot
    \\
    &\sqrt{d\vert S\vert \vert A\vert \vert T\vert \log(1+\frac{T}{d\lambda})}.
\end{align*}
\end{lemma}
\begin{proof}
    \begin{align*}
    \sum_{m=1}^T  &\sum_{h=1}^{H^m} \sum_{s' \in S}  
    \widetilde{\mathcal{V}}^{\pi}_{\mathcal{M}(c_k)} (s') \cdot
    \\
    &\bigg( P^{c_m}(s' \mid s_h^m,a_h^m) -  \widetilde{P}^{c_m} (s' \mid s_h^m,a_h^m) \bigg)
    \\
    &\leq
    \costbound \sum_{s \in S}\sum_{a \in A}\sum_{m=1}^M n_m(s_h^m,a_h^m) \cdot
    \\
    &\quad \Vert P^{c_m}(\cdot \mid s_h^m,a_h^m)
    - \widetilde{P}^{c_m}(\cdot \mid s_h^m,a_h^m) \Vert_1
    \tag{Hölder's inequality}
    \\
    &\leq
    \costbound \sum_{s \in S}\sum_{a \in A}\sum_{m=1}^M n_m(s_h^m,a_h^m) \cdot
    \\
    &\Vert c_m\Vert_{ \bar V_m^{-1} (s,a)}
    \| P^\star(s_h^m,a_h^m)
    - \widetilde{P}_m(s^m_h,a^m_h)\|_{\bar V_m(s,a)}
    \tag{Hölder's inequality}
    \\
    &\leq
    \costbound \sum_{s \in S}\sum_{a \in A}\sum_{m=1}^M n_m(s_h^m,a_h^m)\cdot
    \\
    &\qquad \Vert c_m\Vert_{\bar V^{-1}_m(s, a)}
    \Bigg(\sqrt{d\log{\frac{8|S|^2 |A| (1+\tau_{s,a}/\lambda)}{\delta}}}
    \\
    &\qquad \qquad + \sqrt{\lambda}\Bigg)
    \tag{\cref{thm:confidenceEllipsoid}}
    \\ 
    &\leq  \costbound \Bigg(\sqrt{d\log{\frac{8|S|^2 |A| (1+T/\lambda)}{\delta}}} +\sqrt{\lambda}\Bigg) \cdot
    \\
    \sum_{s \in S}
    &\sum_{a \in A}
    \Bigg[\sum_{m=1}^M n_m(s,a)
    \Vert c_m\Vert_{\bar{V}^{-1}_m(s,a)}\Bigg]
    \\
    &\le
    \costbound \Bigg(
        \sqrt{d\log{\frac{8 |S|^2 |A| (1+T/\lambda)}{\delta}}} +\sqrt{\lambda}
        \Bigg) \cdot
        \\
        &\sqrt{d\vert S\vert \vert A\vert  T \log(1+\frac{T}{d\lambda})},
    \end{align*}
    where the last equality follows by~\cref{lemma:decomposition-aux} in \cref{sec:r-bound}.
\end{proof}
\cref{eq:hoff-reg-decomp-3-main} measures the difference between observed value at the next state and the expected value under the transition model. It is given directly from the HPE, and its proof is straightforward from Azuma's inequality and supplied in~\cref{lem:azuma's-bounds}.


\cref{eq:hoff-reg-decomp-4-main} represents the the error in estimating the loss function, and is proved in a similar way to~\cref{lem:sum-conf-bounds-main}.


Finally, \cref{thm:regret-bound-main} is the combination of~\cref{lem:hoff-switching-cost,lem:sum-conf-bounds-main,lem:azuma's-bounds,lem:cost-bound-regret} when placing the bound on $T$ and $M$ from \cref{lemma:number-steps-bound-main}. 
The complete proof is found in \cref{sec:r-bound}.
\section{Acknowledgements}
 This project is supported by the Israel Science Foundation (ISF, grant number 2250/22).

\bibliographystyle{plainnat}
\bibliography{references}





\clearpage
\appendix
\thispagestyle{empty}

\onecolumn













\section{Proofs}
\subsection{Proof of \texorpdfstring{\cref{lemma:number-steps-bound-main}}{}}\label{sec:t-bound}

\begin{paragraph}{Lemma}(Restatement of \cref{lemma:number-steps-bound-main})
    Assume \(\lambda \ge 1\). Then, \(\Omega\) holds With probability at least \(1-\delta\). 
This implies that the total number of steps of the algorithm is:
\[
    T
    =
    \widetilde{O}
    \Bigg(
        \frac{K\costbound}{\ell_{\min}} + \frac{|S|^3|A|{d^{2} B^3_*}}{\ell_{\min}^3}\log^{2}{\frac{1}{\delta}}
    \Bigg).          
\]
\end{paragraph}

\begin{lemma}\label{lem:opt-val-bound}
Let \(m\) be an interval. If $\Omega$ holds, then $\optimisticctg{m}(s) \leq \ctg{\pi^\star}(s) \leq \costbound$ for every $s \in S$.
\end{lemma}
\begin{proof}
Since $\Omega$ holds, Equations~\ref{eq:loss-conf-set} and~\ref{eq:dynamics-conf-set} hold, which implies that the true dynamics $P^\star$ and the true loss $L^\star$ where considered in the minimization. Therefore, the chosen value function in~\cref{ln:optimism} in~\cref{alg:alg} must be lower then that of $P^\star$ and $L^\star$, $\ctg{\pi^\star}(s) \leq \costbound$ for every $s \in S$.
\end{proof}

\begin{lemma}
    \label{lem:close-mdps-proper}
    Let \(\widetilde{\pi}\) be a policy, \(\widetilde{P}\) be a transition function and \(\widetilde{\ell}\) a loss function. Denote the loss-to-go of \(\widetilde{\pi}\) with respect to \(\widetilde{P}\) by \(\widetilde{\mathcal{V}}\).
    Assume that for every \(s \in S\), \(\widetilde{\mathcal{V}} (s) \leq \costbound\) and 
    \begin{equation}\label{eq:dynamics-approx-assumption}
        \bigl\lVert \widetilde P(\cdot \mid s,\Tilde{\pi}(s)) 
        - 
        P(\cdot \mid s,\Tilde{\pi}(s)) \bigr\rVert_1 
        \leq 
        \frac{\ell(s,\Tilde{\pi}(s))}{2 \costbound}.
    \end{equation}
    
    \begin{equation}\label{eq:loss-approx-assumption}
        |\widetilde{\ell}(s,\Tilde{\pi}(s)) 
        - 
        \ell(s,\Tilde{\pi}(s))|
        \leq 
        \frac{\ell(s,\Tilde{\pi}(s))}{4}.
    \end{equation}
    
    Then, \(\Tilde{\pi}\) is proper (with respect to \(P\)), and it holds that \( \ctg{\tilde{\pi}} (s) \leq 4 \costbound\) for every \(s \in S\).
\end{lemma}

\begin{proof}
Consider the Bellman  equations of \(\Tilde{\pi}\) with respect to transition function \(\widetilde{P}\) and the loss $\widetilde \ell(s,a)$ at some state \(s \in S\) (see \cref{lem:bertsekas-proper}) defined as:

\begin{align}
    \nonumber
    \optimisticctg{} (s) 
    &= 
    \widetilde{\ell}(s,\Tilde{\pi}(s)) + \sum_{s' \in S} \widetilde{P} (s' \mid s,\tilde{\pi}(s)) \optimisticctg{}(s') \\
    \label{eq:opt-bell}
    &=
    \widetilde{\ell}(s,\Tilde{\pi}(s)) + \sum_{s' \in S} P (s' \mid s,\tilde{\pi}(s)) \optimisticctg{}(s')
    +
    \sum_{s' \in S} \optimisticctg{}(s') \left( \widetilde{P} (s' \mid s,\tilde{\pi}(s)) - P(s' \mid s,\tilde{\pi}(s)) \right).
\end{align}

Notice that by our assumptions and using H\"{o}lder inequality,
\begin{align*}
    \left| \sum_{s' \in S} \optimisticctg{}(s') \left( \widetilde{P} (s' \mid s,\tilde{\pi}(s)) - P(s' \mid s,\tilde{\pi}(s)) \right) \right| 
    & \leq 
    \lVert \widetilde{P} (\cdot \mid s,\Tilde{\pi}(s)) - P(\cdot \mid s,\Tilde{\pi}(s)) \rVert_1 \cdot \lVert \optimisticctg{} \rVert_\infty \\
    & \leq 
    \frac{\ell(s,\Tilde{\pi}(s))}{2 \costbound} \cdot \costbound 
    = 
    \frac{\ell(s,\Tilde{\pi}(s))}{2}\ .
\end{align*}

Plugging this into Eq. \ref{eq:opt-bell}, we obtain
\begin{align*}
    \optimisticctg{} (s) 
    &\geq 
    \widetilde{\ell}(s,\Tilde{\pi}(s)) + \sum_{s' \in S}  P (s' \mid s,\tilde{\pi}(s)) \optimisticctg{}(s')
    -
    \frac{\ell(s,\Tilde{\pi}(s))}{2} \\
    &= 
    \widetilde{\ell}(s,\Tilde{\pi}(s))  - 
    \frac{\ell(s,\Tilde{\pi}(s))}{2} + \sum_{s' \in S}  P (s' \mid s,\tilde{\pi}(s)) \optimisticctg{}(s')\\
    &\geq
     - |\widetilde{\ell}(s,\Tilde{\pi}(s))  - 
    \ell(s,\Tilde{\pi}(s))|  + 
    \frac{\ell(s,\Tilde{\pi}(s))}{2} + \sum_{s' \in S}  P (s' \mid s,\tilde{\pi}(s)) \optimisticctg{}(s') \\
    &\geq\tag{\text{\cref{eq:loss-approx-assumption}}} - \frac{\ell(s,\Tilde{\pi}(s))}{4}
    + 
    \frac{\ell(s,\Tilde{\pi}(s))}{2} + \sum_{s' \in S}  P (s' \mid s,\tilde{\pi}(s)) \optimisticctg{}(s')\\ 
    &=\frac{\ell(s,\Tilde{\pi}(s))}{4} + \sum_{s' \in S}  P (s' \mid s,\tilde{\pi}(s)) \optimisticctg{}(s').
\end{align*}

Therefore, defining \({\ctg{}}' = 4\optimisticctg{}\),
then \({\ctg{}}'(s) \geq \ell(s,\Tilde{\pi}(s)) + \sum_{s' \in S} P (s' \mid s,\tilde{\pi}(s)) {\ctg{}}'(s')\) for all \(s \in S\). 
The statement now follows by \cref{lem:bertsekas-proper}.
\end{proof}

\begin{lemma}
    \label{lemma:dynamics-bound}
    Suppose that state-action pair $s,a$ was visited $\tau_{s,a}$ times during interval $m$. Then:
    \begin{align*}
            &\mathrm{tr} 
            \Bigg( 
                (P^\star(s,a) -\hat{P}_m(s,a)')\bar{V}_{\tau_{s,a}}(P^\star(s,a)-\hat{P}_m(s,a)')^\top
            \Bigg) \\
            &\qquad \leq 
            |S| 
            \Bigg(
                {\sqrt{d \log{\frac{8|S|^2|A|(1 + \tau_{s,a})/\lambda)}{\delta}}} +\sqrt{\lambda}}
            \Bigg)^2,
    \end{align*}
    and the confidence sets of the dynamics and loss, defined in \cref{eq:loss-conf-set,eq:dynamics-conf-set} contain $P^\star(s,a)$, $L^\star(s,a)$ for all intervals $m$ simultaneously with probability at least $1-\frac{\delta}{4}$.
\end{lemma}

\begin{proof}
    by Line~\ref{ln:loss-dynamics-ls} in \cref{alg:alg} and~\cref{eq:dynamics-ls}: 
    \begin{align*}
        \hat{P}_m(s,a)' 
        &=
        \argmin_{P \in \mathbb{R}^{|S| \times d}} \sum_ {(s', c) \in D(s,a)} \|P c - e_{s'}\|_2^2 + \lambda \|P\|_F^2
        \\
        &=
        \argmin_{P \in \mathbb{R}^{|S| \times d}} \sum_ {(s', c) \in D(s,a)} \sum_{s'' \in S}\|P_{s''} c - e_{s'}(s'')\|_2^2 + \lambda\sum_{s'' \in S}\|P_{s''}\|^2
        \\
        &=
        \argmin_{P \in \mathbb{R}^{|S| \times d}} \sum_{s'' \in S}
        \brk*{\sum_ {(s', c) \in D(s,a)}\|P_{s''} c - e_{s'}(s'')\|_2^2 + \lambda\|P_{s''}\|^2},
    \end{align*}
    where \(P_{s''}\) is the row of $P$ corresponding to state $s''$.
    Therefore, we can solve for each \(P_{s''}\) separately:
    \begin{align*}
        \hat{P}_m(s'' \mid s,a)' 
        &= \argmin_{P_{s''} \in \mathbb{R}^d} \sum_ {(s', c) \in D(s,a)} \|P_{s''} c - e_{s'}(s'')\|_2^2 + \lambda \|P_{s''}\|^2.
    \end{align*}
    


    by~\cref{thm:confidenceEllipsoid} we have with probability at least $1-\delta'$ where $\delta' =\frac{\delta}{8|S|^2 |A|}$ for each $(s',a,s)$ of $P^\star(s'\mid s,a)$:
    \begin{align*}
        \Vert P^\star(s' \mid s,a)-\hat{P}_m(s' \mid s,a)' \Vert_{\bar{V}_m{(s,a)}} \leq \sqrt{d\log{\frac{1+ t/\lambda}{\delta'}}} +\sqrt{\lambda},
    \end{align*}
    
    Now aggregate all rows to the matrix $\hat{P}(s,a)' \in \mathbb{R}^{| S| \times d}$ by the union bound. By \cref{lem:dynamics-proj-main} we have that the distance is not increased:
    \begin{align*}
        \|P^\star(s,a)-\hat P_m(s,a)\|_{\bar V_m(s,a)}^2
        \le
        |S| \brk*{\sqrt{d\log{\frac{8|S|^2 |A| (1+t/\lambda)}{\delta}}} +\sqrt{\lambda}}^2.
    \end{align*}
    Finally, use the union bound for all $s, a$ together with the loss for all $s,a$ $(\delta'' =\frac{\delta}{8|S| |A|})$.
\end{proof}

\begin{lemma}
    \label{lem:known-state-dynamics}
    Let \(m\) be an interval and suppose that $\Omega$ holds. Denote the context at interval $m$ as $c$. 
    If \((s,a)\) are a known state-action pair, then: 
    \[
        \lVert \hat{P}_m(\cdot \mid s,a)\cdot c - P^\star(\cdot \mid s,a)\cdot c \rVert_1 
        \leq 
        \frac{\ell(s,a)}{2 \costbound}\ .
    \]
\end{lemma}
\begin{proof}
    From \cref{lem:dynamics-proj-main} we have with probability at least $1-\delta$:
    \begin{align*}
        &\mathrm{tr}\brk{{(P^\star(s,a)-\hat{P}_m(s,a))\bar{V}_m{(s,a)}(P^\star(s,a) - \hat{P}(s,a))^\top}}
        \\
        &\qquad \le
        \mathrm{tr}\brk{{(P^\star(s,a)-\hat{P}_m(s,a)')\bar{V}_m{(s,a)}(P^\star(s,a) - \hat{P}_m(s,a)')^\top}} \\
        &\qquad \leq 
        |S| \brk*{\sqrt{d\log{\frac{8|S|^2 |A|(1+\tau_{s,a}/\lambda)}{\delta}}} +\sqrt{\lambda}}^2.
    \end{align*}
    Therefore:
    \begin{align}
        \Vert{P^\star(s,a) \cdot c_m-\hat{P}_m(s,a) \cdot c_m}\Vert_1
        &= 
        \sum_{s' \in S} \abs{P^\star(s'\mid s,a)\cdot c_m - \hat{P}_m(s' \mid s,a)\cdot c_m }
        \nonumber \\
        &\leq
        \sum_{s' \in S}\Vert P^\star(s'\mid s,a)-\hat{P}_m(s' \mid s,a)\Vert_{\bar{V}_m{(s,a)}}
        \Vert c_m\Vert_{\bar{V}_m{(s,a)}^{-1}}  \tag{H\"{o}lder's inequality} \\
        &\leq
        \Vert c_m\Vert_{\bar{V}_m{(s,a)}^{-1}}\cdot |S| \brk*{\sqrt{d\log{\frac{8|S|^2 |A| (1+\frac{\tau_{s,a}}{\lambda})}{\delta}}} +\sqrt{\lambda}}.
        \label{eq:temp1}
    \end{align}
    
    Since the pair $(s,a)$ is known, we have:
\begin{align}\label{eq:temp2}
    \Vert c_{m}\Vert_{\bar{V}_m(s,a)^{-1}} < \frac{\ell_{\min}}{10 \costbound \max\brk[c]*{|S|(\sqrt{d\log{\frac{8|S|^2 |A|(1 + \tau_{s,a}/\lambda)}{\delta}}} + \sqrt{\lambda}),\sqrt{\log(4m/\delta)}}}.
\end{align}
    
    Injecting \cref{eq:temp2} to \cref{eq:temp1}:
    \begin{align*}
    \Vert{P_m^\star(s,a)\cdot c_m-\hat{P}_m(s,a)\cdot c_m}\Vert_1 \leq \frac{\ell_{\min}}{10\costbound} \leq 
    \frac{\ell(s,a)}{2 \costbound}.
    \qquad \qedhere
    \end{align*}
\end{proof}

\begin{lemma}\label{lem:known-state-loss}
    Let \(m\) be an interval in episode with context \(c\), and suppose that $\Omega$ holds. 
    If \((s,a)\) is a known state-action pair, then: \[
        \abs{{\ell}^\star(s,a) \cdot c_m - \hat{\ell}_m(s,a) \cdot c_m} 
        \leq
        \frac{\ell(s,a)}{4}. 
    \]
\end{lemma}

\begin{proof}
    Following Line~\ref{ln:loss-dynamics-ls} in \cref{alg:alg} and~\cref{eq:loss-ls}.
    we apply \cref{thm:confidenceEllipsoid}) to get for probability at least $1-\delta$:
\begin{align*}
    \Vert \ell^\star(s,a) -\hat{\ell}_m(s,a)\Vert_{\bar{V}_m{(s,a)}} \leq \sqrt{d\log{\frac{|S|^2|A|(1+\frac{\tau_{s,a}}{\lambda})}{\delta}}} +\sqrt{\lambda}.
\end{align*}

Thus, we have:
\begin{align}
    \abs{\ell^\star(s,a)\cdot c - \hat{\ell}(s,a)\cdot c}
    \nonumber
    &\underset{\text{\{i\}}}{\leq}
    \Vert \ell^\star(s,a) - \hat{\ell}_i(s,a)\Vert_{\bar{V}_m{(s,a)}}
    \Vert c\Vert_{\bar{V}_m{(s,a)}^{-1}} \\
    \label{eq:lemma-loss-bound}
    &=
    \Vert c\Vert_{\bar{V}_m^{-1}(s,a)} \cdot \Bigg(\sqrt{d\log{\frac{8|S|^2 |A| (1+ \frac{\tau_{s,a}}{\lambda})}{\delta}}} +\sqrt{\lambda}\Bigg) \\
    &\underset{\text{\{ii\}}}{\leq}
    \Vert c\Vert_{\bar{V}_m^{-1}(s,a)} \cdot \Bigg(\sqrt{d\log{\frac{8|S|^2 |A| (1+\frac{T}{\lambda})}{\delta}}} +\sqrt{\lambda}\Bigg),
    \nonumber
\end{align}
where \{i\} follows from Hölder's inequality and \{ii\} because \(t \leq T\).
Now inject~\cref{eq:ineq-known}, since the pair (s,a) is known, to~\cref{eq:lemma-loss-bound}.
%

\begin{align*}
|{\ell^\star(s,a)\cdot c - \hat{\ell}_m(s,a)\cdot c}| \leq \frac{\ell_{\min}}{10} \leq 
\frac{\ell(s,a)}{4}.
\qquad \qedhere
\end{align*}
\end{proof}

\begin{lemma}\label{lem:aux-bound-2B-known}
    Assume \(\lambda \ge 1\).
    Let $n_m(s,a)$ be the number of visits to $s,a$ during interval $m$. Then:
        \[
            \sum_{m=1}^{M} n_m(s,a) \| c_m\|^2_{\bar{V}_m^{-1}(s,a)}
            \leq
            d\log(1+\frac{\tau_{s,a}}{\lambda d}).
        \]
\end{lemma}
\begin{proof}
    Let \(c_m\) be the context associated with the interval \(m\) and let \(\bar V_m\) be the matrix \(V_{\tau_{s,a}}\) that is set at the beginning of the interval \(m\).
    \begin{align*}
        \sum_{m=1}^{M} n_m(s,a) \| c_m\|^2_{\bar{V}_m(s,a)^{-1}}
        &=
        \sum_{m=1}^{M} n_m(s,a) c_m^\top \bar{V}^{-1}_m(s,a) c_m \\
        &=
        \sum_{m=1}^{M} \mathrm{tr}(\bar{V}_m^{-1}(s,a) n_m(s,a) c_m c_m^\top) \\
        &=
        \tag{\(\bar{V}_{m+1}(s,a)=\bar V_m(s,a) + n_m(s,a) c_m c_m^\top\)}
        \sum_{m=1}^{M} \mathrm{tr}(\bar{V}_m(s,a)^{-1}(\bar{V}_{m+1}(s,a) - \bar{V}_m(s,a)) \\
        &\leq
        2 \sum_{m=1}^{M} (\log{\det(\bar{V}_{m+1}(s,a))} - \log{\det(\bar{V}_m(s,a))})  
        \tag{*} \\
        &=
        2(\log{\det(\bar{V}_{M+1}(s,a))} - \log{\det(\bar{V}_1(s,a))}) 
        \tag{Telescopic sum} \\
        &= 
    2(\log{\det(\lambda^{-1}\bar{V}_{M+1}(s,a)}),
    \end{align*}
    where (*) follows due to concavity of $\log\det$, and the fact that
    $
        n_m(s,a) c_m^\top \bar V_m^{-1}(s,a) c_m
        \le 1,
    $
    that holds because of the following.
    If $(s,a)$ is unknown states, $n_m(s,a) \leq 1$, and $c_m^\top \bar V_m^{-1} c_m \le \|c_m\|^2 / \lambda \le 1$, since $\lambda \ge 1$ and $\|c_m\| \le 1$ by assumption. 
    For known $(s,a)$, by \cref{item:interval-cost-bound} of the HPE together with \cref{eq:ineq-known} yields 
    \[
        n_m(s,a) c_m^\top \bar V_m^{-1} c_m
        \le 
        \frac{48\costbound \log(4m/\delta)}{\ell_{\min}} \cdot \frac{\ell^2_{\min}}{100\costbound^2 \log(4m/\delta)}
        \le 1,
    \]
    since $\ell_{\min} \le 1$ and $\costbound \ge 1$ by assumption.
    This yields inequality $(i)$ of the Lemma. 
    
    Next,
    \begin{align*}
        \log{\det(\lambda^{-1}\bar{V}_{M+1}(s,a))}
        &\leq\tag{AM-GM}
        \log \text{tr}\brk*{\frac{1}{\lambda d}\bar{V}_{M+1}(s,a)}^d \\
        &=
        d\log \text{tr} \left( {\frac{1}{\lambda d}(\lambda I + \sum_{t=1}^{\tau_{s,a}}c_tc_t^T)} \right)  \\
        &=
        d\log(1+\frac{1}{\lambda d}\sum_{t=1}^{\tau_{s,a}}\Vert c_t\Vert^2) \\
        &\leq
        d\log(1+\frac{\tau_{s,a}}{\lambda d}).
        \qedhere
    \end{align*}
\end{proof}
    
    

\begin{lemma}[\citealp{pmlr-v119-rosenberg20a}]\label{lem:proper-not-run-log}
Let \(\pi\) be a proper policy such that for some \(v > 0\), \(\ctg{\pi} (s) \leq v\) for every \(s \in S\).  
Then, the probability that the loss of \(\pi\) to reach the goal state from any state \(s\) is more than \(m\), is at most \(2e^{-m/4v}\) for all \(m \ge 0\).
Note that a total loss of at most $m$ implies that the number of steps is bounded above by \(\frac{m}{\ell_{\min}}\).
\end{lemma}

Let us denote the trajectory visited in interval \(m\) by $U^m = ( s_1^m, a_1^m , \ldots , s_{H^m}^m, a_{H^m}^m, s_{H^m+1}^m)$ where $a_h^m$ is the action taken in $s_h^m$, and $H^m$ is the length of the interval.
In addition, the concatenation of the trajectories in the intervals up to and including interval $m$ is denoted by $\bar U^{m} = \cup_{m'=1}^m U^{m'}$.

\begin{lemma}\label{lemma:interval-cost-bound}
    Let \(m\) be an interval during an episode associated with \(c_k\).
    For all \(r \ge 0\), we have \[
        \Pr \Biggl[\sum_{h=1}^{H^m} \ell \bigl(s_h^m, a_h^m \bigr) > r \mid \bar U^{m-1} \Biggr] 
        \le 
        3e^{-r/16 \costbound}.
    \]
\end{lemma}

\begin{proof}
    For a given SSP $\cmdp{c_k}$ define the contracted SSP $\cmdpk{c_k} = (S^\text{know}, A, P^\text{know}_{c_k}, \ell_{c_k}, \sinit^{c_k})$ in which every state $s \in S$ such that  $(s,\tilde{\pi}^m(s))$ is unknown is contracted into  the goal state. 
    Let $\cmdp{c_k}, \cmdpkt{c_k}$ be the contracted SSP's corresponding to $P(\cdot)$ and $\widetilde{P}(\cdot)$ respectively;  
    Let $\ctg{m}_\text{know}$ and $\optimisticctg{m}_\text{know}$ be the cost-to-go of $\tilde{\pi}^m$ in $\cmdpk{c_k}$ and $\cmdpkt{c_k}$, respectively. 
    Now apply \cref{lem:close-mdps-proper} in $\cmdpk{c_k}$ and obtain that $\ctg{m}_{\text{know}} (s) \leq 4 \costbound$ for every $s \in S^\text{know}$.

    Notice that reaching the goal state in $\cmdpk{c_k}$ is equivalent to reaching the goal state or an unknown state-action pair in $M$, and also recall that all state-action pairs in the interval are known except for the first one.
    Thus, from \cref{lem:proper-not-run-log},
    \begin{align*}
        \mathbb{P} \Biggl[\sum_{h=2}^{H^m} \ell \bigl(s_h^m, a_h^m \bigr) \indgeventi{m} > r-1 \mid \bar U^{m-1} \Biggr]
        \le
        2e^{-(r-1)/16\costbound}
        <
        3e^{-r / 16 \costbound}.
    \end{align*}
    
    Where the last inequality is because $\costbound \geq1$. 
    Since the first state-action in the interval might not be known, its loss is at most \(1\), and therefore
    \begin{align*}
        \mathbb{P} 
        \Biggl[
            \sum_{h=1}^{H^m} \ell \bigl(s_h^m, a_h^m \bigr) \indgeventi{m} > r \mid \bar U^{m-1} 
        \Biggr]
        &\le
        \mathbb{P} 
        \Biggl[
            \sum_{h=2}^{H^m} \ell \bigl(s_h^m, a_h^m \bigr) \indgeventi{m} > r-1 \mid \bar U^{m-1} 
        \Biggr]
        \\
        &\le
        3 e^{-r / 16 \costbound}. \qedhere
    \end{align*}
\end{proof}

\begin{lemma}\label{lem:hpe-interval-loss-aux}
    $\sum_{h=1}^{H_m} \ell(s_h^m,a_h^m) \le \highprobcostbound{m}$ for all intervals \(m\) simultaneously with probability at least $1 - \frac{\delta}{4}$.
\end{lemma}
\begin{proof}
   From Lemma~\ref{lemma:interval-cost-bound}, we have:
    \begin{equation}\label{eq:interval-cost-bound}
                \mathbb{P} \Biggl[\sum_{h=1}^{H^m} \ell \bigl(s_h^m, a_h^m \bigr) \indgeventi{m} \leq
        48 \costbound \log{\frac{4m}{\delta}}
        \mid \bar U^{m-1} \Biggr] 
        \ge 
        1-\frac{\delta}{16m^2}.
    \end{equation}   
    By the union bound \cref{eq:interval-cost-bound} holds for all intervals \(m\) simultaneously with cumulative probability of at least \(1 - \frac{\delta}{4}\).
\end{proof}
\begin{paragraph}{Lemma}\label{lem:hpe-interval-loss} (Restatement of \cref{lem:hpe-interval-loss-main})
The HPE holds with probability at least $1-\delta$.
\end{paragraph}
\begin{proof}
  Apply \cref{lemma:dynamics-bound,lem:hpe-interval-loss-aux,lem:azuma's-bounds} and the union bound.
\end{proof}

Now, we are ready to conclude the proof of Lemma ~\ref{lemma:number-steps-bound-main}.

\begin{proof}[Proof of \cref{lemma:number-steps-bound-main}]
    \label{proof:number-steps-bound}
    \cref{lem:hpe-interval-loss} guarantees that $\Omega$ holds for all \(m\) with probability at least \(1-\delta\).
    \cref{lem:bound-2B-known} bounds the number of times the pair $(s,a)$ is unknown. 
    Since $\Omega$ holds, the combination of \cref{lem:close-mdps-proper,lem:known-state-dynamics,lem:known-state-loss} ensures that the policy calculated in~\cref{ln:optimism} in~\cref{alg:alg} is proper and therefore reaches the goal state or an unknown state-action pair with probability 1.
    If the cumulative loss of all intervals is bounded by the HPE (and therefore so is the length of the interval), we can use \cref{obsv:num-intervals} to conclude that:
    \[
       T
       =
        \mathcal{\widetilde{O}}
        \Bigg(
            \frac{K \costbound}{\ell_{\min}} + \frac{|S|^3|A|{d^{2} \costbound^3}}{\ell_{\min}^3}(\log^2{\frac{1}{\delta}} )
        \Bigg)    
        \qedhere
    \]
\end{proof}

\subsection{Proof of Theorem \ref{thm:regret-bound-main}}\label{sec:r-bound}

\paragraph{Theorem}(Restatement of \cref{thm:regret-bound-main}).
\label{thm:regret-bound}
    With probability at least \(1-\delta\) the regret of algorithm~\ref{alg:alg} is bounded as follows:
    \begin{align}
    R_k
        =
        \widetilde{O}
        \Bigg(
            \frac{\costbound^3 d^2|S|^3 |A|}{\ell_{\min}^2}
            \log\frac{1}{\delta}
            +
            \sqrt{K \cdot\frac{d\vert S\vert^3 \vert A\vert \costbound^2 \log (1/\delta)}{\ell_{\min}}}
        \Bigg).
    \end{align}
\begin{lemma}\label{lem:regret-tilde-bound}
    Suppose that the HPE holds, then:
    \begin{align*}
        R_K
        &\le 
        \costbound |S| |A| N_{s,a}
        +
        2\costbound \vert S \vert \Bigg(\sqrt{d\log{\frac{8|S|^2|A|(1+T/\lambda)}{\delta}}} +\sqrt{\lambda}\Bigg)
        \sqrt{d\vert S\vert \vert A\vert \vert T\vert \log(1+\frac{T}{d\lambda})} \\
        &+
        \costbound \sqrt{T \log \frac{4 T}{\delta}}.
    \end{align*}
\end{lemma}
To analyze \(R_{K}\), we begin by plugging in the Bellman optimality equation of \(\Tilde{\pi}_m\) with respect to \(\widetilde{P}_m\) into \(R_{K}\). This allows us to decompose \(R_{K}\)into four terms as follows.
\begin{align}
        R_K
    \label{eq:hoff-reg-decomp-1}
    & = \sum_{k=1}^K \sum_{t=1}^{I^k} 
    \left(
        \widetilde V_m (s_{h}^m) - \widetilde V_m (s_{h+1}^m)
    \right)
    -
    \sum_{k=1}^K \min_{\pi \in \propset{c_k}} \widetilde V_m(\sinit)
    \\
    & = \sum_{m=1}^M \sum_{h=1}^{H^m} 
    \left(
        \widetilde V_m (s_{h}^m) - \widetilde V_m (s_{h+1}^m)
    \right)
    -
    \sum_{k=1}^K \min_{\pi \in \propset{c_k}} \widetilde V_m(\sinit)
    \\
    \label{eq:hoff-reg-decomp-2}
    & \qquad +
    \sum_{m=1}^M  \sum_{h=1}^{H^m}  \sum_{s' \in S}
    \widetilde V_m (s'_m)
    \left(
        P^{c_k}(s' \mid s_h^m,a_h^m) -  \widetilde{P}^{c_k}_m(s' \mid s_h^m,a_h^m) 
    \right)
    \\
    \label{eq:hoff-reg-decomp-3}
    & \qquad + 
    \sum_{m=1}^M
    \left( \sum_{h=1}^{H^m}
        \widetilde V_m (s_{h+1}^m)
        -
        \sum_{s' \in S} P^{c_k}(s' \mid s_h^m,a_h^m)
        \widetilde V_m (s'_m)
    \right).
    \\
    \label{eq:hoff-reg-decomp-4}
    & \qquad + 
    \sum_{m=1}^M \sum_{h=1}^{H^m} \ell^{c_k}(s,a) - \widetilde{\ell}^{c_k}(s,a).
\end{align}

\begin{lemma}\label{lem:hoff-switching-cost}
    \[
     \sum_{m=1}^M \sum_{h=1}^{H^m} 
    \left(
        \widetilde V_m (s_{h}^m) - \widetilde V_m (s_{h+1}^m)
    )\right)
    \leq
    \costbound |S| |A| \cdot N_{s,a}
    +
    \sum_{k=1}^K \min_{\pi \in \propset{c_k}} \valf{c_k}{\sinit}
    \]
\end{lemma}
\begin{proof}
    Note that per interval 
    \(
 \sum_{m=1}^M \sum_{h=1}^{H^m} 
    \left(
        \widetilde V_m (s_{h}^m) - \widetilde V_m (s_{h+1}^m)
    )\right)
    \)
    is a telescopic sum which equals 
    \(
        \widetilde V_m (s_1^m)
        -
        \widetilde V_m (s_{H^{m+1}}^m)
    \).
    Furthermore, for every two consecutive intervals \(m,m+1\) one of the following occurs:
    \begin{enumerate}[label=(\roman*)]
        \item If interval \(m\) ended in the goal state then
        \(
            \widetilde V_m (s_{h+1}^m) =\widetilde V_m (g) = 0
        \)
        and
        \(
           \widetilde V_m (s_1^{m+1}) = \widetilde V_m (\sinit)
        \)
        Thus,
        \begin{align*}
            \widetilde V_m (s_1^{m+1})  
            - 
            \widetilde V_m (s_{H^m + 1}^m)
            & = 
            \widetilde V_m (s_1^{m+1}) \\
            &= \widetilde V_m (\sinit) \\
            &\leq
            \min_{\pi \in \propset{c_k}} V^{\pi}_{M(c_k)} (\sinit),
            \tag{\cref{lem:opt-val-bound}}
        \end{align*}
        which adds up to 
        \(
            \sum_{k=1}^K \min_{\pi \in \propset{c_k}} \widetilde V_m(\sinit)
        \).
    
    \item If interval \(m\) ended in an unknown state then
    \[
        \widetilde V_m (s_1^{m+1})  
        - 
        \widetilde V_m (s_{H^m + 1}^m)
        \leq 
        \widetilde V_m (s_1^{m+1})
        \leq 
        \costbound.
    \]
    This happens at most \(|S| |A| N_{s,a}\) times.
    \qedhere
\end{enumerate}
\end{proof}

\begin{lemma}\label{lemma:decomposition-aux}
\begin{align*}
    \sum_{s,a}\sum_{m=1}n_m(s,a)\Vert c_m\Vert_{\bar{V}^{-1}_m} \leq 
    \sqrt{d\vert S\vert \vert A\vert \vert T\vert \log(1+\frac{T}{d\lambda})}.
\end{align*}
\end{lemma}
\begin{proof}
\begin{align*}
    \sum_{m=1}n_m(s,a)\Vert c_m\Vert_{\bar{V}^{-1}_m(s,a)} 
    &= 
    \sum_{m=1}\sqrt{n_m(s,a)}\sqrt{n_m(s,a)}\Vert c_m\Vert_{\bar{V}^{-1}_m(s,a)} \\
    &\leq\tag{Cauchy-Schwartz}
    \sqrt{\sum_{m=1}n_m(s,a)}\sqrt{\sum_{m=1}n_m(s,a)
    \cdot\Vert c_m\Vert^2_{\bar{V}^{-1}_m(s,a)}} \\
    &\leq\tag{\cref{lem:bound-2B-known}}
    \sqrt{\tau_{s,a}} \sqrt{d\log\brk*{1+\frac{\tau_{s,a}}{d\lambda}}}.
\end{align*}
Therefore, 
\begin{align*}
    \sum_{s,a}\sum_{m=1}n_m(s,a)\Vert c_m\Vert_{\bar{V}^{-1}_m(s,a)}
    \leq 
    \sum_{s,a}\sqrt{d\log(1+\frac{T}{d\lambda})}\sqrt{\tau_{s,a}}  
    \leq
    \sqrt{d\vert S\vert \vert A\vert T \log(1+\frac{T}{d\lambda})}.
\end{align*}
Where the last inequality holds because we can apply Cauchy-Schwartz inequality on $\sum_{s,a}\sqrt{d\log(1+\frac{T}{d\lambda})}$ is fixed for all $s,t$ and because $\sum_{s,a}\tau_{s,a}=T$.
\end{proof}
%

\begin{lemma} 
\label{lem:azuma's-bounds}
With probability at least \(1 - \delta / 2\),
\begin{equation}
            \sum_{m=1}^M
                \left( \sum_{h=1}^{H^m}
                \widetilde V_m(s_{h+1})-
            \sum_{s' \in S} P^{c_k}(s' \mid s_h^m,a_h^m)
                \widetilde V_m(s')
            \right)
            \leq 
            \costbound \sqrt{T \log \frac{4 T}{\delta}}.
\end{equation}
\end{lemma}

\begin{proof}
Consider the infinite sequence of random variables
\[
    X_t
    = 
    \left(
        \sum_{h=1}^{H^m}
            \widetilde V_m(s_{h+1})
            -
            \sum_{s' \in S} P^{c_k}(s' \mid s_h^m,a_h^m) \widetilde V_m(s')
    \right),
\]
where \(m\) is the interval containing time \(t\), and \(h\) is the index of time step \(t\) within  interval \(m\).
Notice that this is a martingale difference sequence, and \(|X_t| \leq \costbound\) by \cref{lem:opt-val-bound}.
Now, we apply anytime Azuma's inequality (\cref{thm:azuma}) to any prefix of the sequence \(\{ X_t \}_{t=1}^\infty\). Thus, with probability at least \(1 - \delta / 2\), for every \(T\):
\[
    \sum_{t=1}^T X_t
    \leq 
    \costbound \sqrt{T \log \frac{4 T}{\delta}}. \qedhere
\]
\end{proof}
Lastly, the following Lemma bounds the difference between the estimated loss to the true loss, reflected in Equation~\ref{eq:hoff-reg-decomp-4}:

\begin{lemma}\label{lem:cost-bound-regret}
Suppose that $\Omega$ holds. It holds than that:
\[
    \sum_{m=1}^M \sum_{h=1}^{H^m} \ell^{c_k}(s,a) - \widetilde{\ell}^{c_k}(s,a)
    \leq
    \Bigg(\sqrt{d\log{\frac{8|S| |A| (1+T/\lambda)}{\delta}}} +\sqrt{\lambda}\Bigg) 
    \sqrt{d\vert S\vert \vert A\vert \vert T\vert \log(1+\frac{T}{d\lambda})}
\]
\end{lemma}

\begin{proof}
\begin{align*}
    &\sum_{m=1}^M \sum_{h=1}^{H^m} \ell^{c_k}(s,a) - \widetilde{\ell}^{c_k}(s,a)
    \\
    &\qquad
    \leq
    \sum_{s \in S}\sum_{a \in A}\sum_{m=1}^M n_m(s,a)\Vert c_t\Vert_{\bar{V}^{-1}_{\tau_{s,a}}}
    \Vert \ell^{c_k}(s,a) - \widetilde{\ell}^{c_k}(s,a)\Vert_{\bar{V}_{\tau_{s,a}}} 
    \\
    &\qquad
    \leq
    \sum_{s \in S}\sum_{a \in A}\sum_{m=1}^M n_m(s,a)\Vert c_t\Vert_{\bar{V}^{-1}_{\tau_{s,a}}}
    \Bigg(\sqrt{d\log{\frac{8|S| |A| (1+{\tau_{s,a}}\lambda)}{\delta}}} +\sqrt{\lambda}\Bigg)
    \\
    &\qquad
    \leq
    \Bigg(\sqrt{d\log{\frac{8|S| |A| (1+T/\lambda)}{\delta}}} +\sqrt{\lambda}\Bigg) 
    \Bigg(\sum_{s, a}\sum_{m=1}^M n_m(s,a)\Vert c_t\Vert_{\bar{V}^{-1}_{\tau_{s,a}}} \Bigg)
\end{align*}
Using \cref{lemma:decomposition-aux}, we conclude:
\begin{align*}
    \sum_{m=1}^M \sum_{h=1}^{H^m} \ell^{c_k}(s,a) - \widetilde{\ell}^{c_k}(s,a) 
    \leq
    \Bigg(\sqrt{d\log{\frac{8|S| |A| (1+T/\lambda)}{\delta}}} +\sqrt{\lambda}\Bigg) 
    \sqrt{d\vert S\vert \vert A\vert \vert T\vert \log(1+\frac{T}{d\lambda})}. 
\end{align*}
\end{proof}

\begin{proof}[Proof of \cref{thm:regret-bound-main}]
    \cref{lem:hoff-switching-cost}~\cref{lem:sum-conf-bounds-main},~\cref{lem:azuma's-bounds},~\cref{lem:cost-bound-regret} and the fact that $\costbound\geq1$ prove \cref{lem:regret-tilde-bound}. 
    ~\ref{lem:hpe-interval-loss} guarantees that the HPE occurs with probability at least $1-\delta$. We conclude the proof by injecting~\cref{lemma:number-steps-bound-main} to \cref{lem:regret-tilde-bound}.
\end{proof}
\section{Concentration inequalities}

\begin{theorem}[Anytime Azuma](~\cite{pmlr-v119-rosenberg20a}) \label{thm:azuma}
    Let \((X_n)_{n=1}^\infty\) be a martingale difference sequence with respect to the filtration \((\mathcal{F})_{n=0}^\infty\) such that \(|X_n| \le B\) almost surely. 
    Then with probability at least \(1-\delta\),
    \[
        \Biggl|\sum_{i=1}^n X_i \Biggr| \le B \sqrt{n \log \frac{2 n}{\delta}}, \qquad \forall n \ge 1.
    \]
\end{theorem}
\newpage
\end{document}